# The Universe of Minds

ROMAN V. YAMPOLSKIY

Computer Engineering and Computer Science
Speed School of Engineering
University of Louisville, USA
roman.yampolskiy@louisville.edu

**Abstract**
The paper attempts to describe the space of possible mind designs by first equating all minds to software. Next it proves some interesting properties of the mind design space such as infinitude of minds, size and representation complexity of minds. A survey of mind design taxonomies is followed by a proposal for a new field of investigation devoted to study of minds, *intellectology*, a list of open problems for this new field is presented.

**Introduction**

In 1984 Aaron Sloman published "The Structure of the Space of Possible Minds" in which he described the task of providing an interdisciplinary description of that structure [1]. He observed that "behaving systems" clearly comprise more than one sort of mind and suggested that virtual machines may be a good theoretical tool for analyzing mind designs. Sloman indicated that there are many discontinuities within the space of minds meaning it is not a continuum, nor is it a dichotomy between things with minds and without minds [1]. Sloman wanted to see two levels of exploration namely: *descriptive* – surveying things different minds can do and *exploratory* – looking at how different virtual machines and their properties may explain results of the descriptive study [1]. Instead of trying to divide the universe into minds and non-minds he hoped to see examination of similarities and differences between systems. In this work we attempt to make another step towards this important goal.

What is a mind? No universal definition exists. Solipsism notwithstanding, humans are said to have a mind. Higher order animals are believed to have one as well and maybe lower level animals and plants or even all life forms. We believe that an artificially intelligent agent such as a robot or a program running on a computer will constitute a mind. Based on analysis of those examples we can conclude that a mind is an instantiated intelligence with a knowledgebase about its environment, and while intelligence itself is not an easy term to define, a recent work of Shane Legg provides a satisfactory, for our purposes, definition [2]. Additionally, some hold a point of view known as Panpsychism, attributing mind like properties to all matter. Without debating this possibility we will limit our analysis to those minds which can actively interact with their environment and other minds. Consequently, we will not devote any time to understanding what a rock is thinking.

If we accept materialism, we have to also accept that accurate software simulations of animal and human minds are possible. Those are known as uploads [3] and they belong to a class comprised of computer programs no different from that to which designed or evolved artificially intelligent software agents would belong. Consequently, we can treat the space of all minds as the space of programs with the specific property of exhibiting intelligence if properly embodied. All programs could be represented as strings of binary numbers, implying that each mind can be represented by a unique number. Interestingly, Nick Bostrom via some thought experiments speculates that perhaps it is possible to instantiate a fractional number of mind, such as .3 mind as opposed to only whole minds [4]. The embodiment requirement is necessary since a string is not a mind, but could be easily satisfied by assuming that a universal Turing machine is available to run any program we are contemplating for inclusion in the space of mind designs. An embodiment does not need to be physical as a mind could be embodied in a virtual environment represented by an avatar [5, 6] and react to simulated sensory environment like a brain-in-a-vat or a "boxed" AI [7].

**Infinitude of Minds**

Two minds identical in terms of the initial design are typically considered to be different if they possess different information. For example, it is generally accepted that identical twins have distinct minds despite exactly the same blueprints for their construction. What makes them different is their individual experiences and knowledge obtained since inception. This implies that minds can't be cloned since different copies would immediately after instantiation start accumulating different experiences and would be as different as two twins.

If we accept that knowledge of a single unique fact distinguishes one mind from another we can prove that the space of minds is infinite. Suppose we have a mind M and it has a favorite number N. A new mind could be created by copying M and replacing its favorite number with a new favorite number N+1. This process could be repeated infinitely giving us an infinite set of unique minds. Given that a string of binary numbers represents an integer we can deduce that the set of mind designs is an infinite and countable set since it is an infinite subset of integers. It is not the same as set of integers since not all integers encode for a mind.

Alternatively, instead of relying on infinitude of knowledgebases to prove infinitude of minds we can rely on the infinitude of designs or embodiments. Infinitude of designs can be proven via inclusion of a time delay after every computational step. Fist mind would have a delay of 1 nano-second, second a delay of 2 nano-seconds and so on to infinity. This would result in an infinite set of different mind designs. Some will be very slow, others super-fast, even if the underlying problem solving abilities are comparable. In the same environment, faster minds would dominate slower minds proportionately to the difference in their speed. A similar proof with respect to the different embodiments could be presented by relying on ever increasing number of sensors or manipulators under control of a particular mind design.

Also, the same mind design in the same embodiment and with the same knowledgebase may in fact effectively correspond to a number of different minds depending on the operating conditions. For example, the same person will act very differently if they are under the influence of an intoxicating substance, under severe stress, pain, sleep or food deprivation, or are experiencing a temporary psychological disorder. Such factors effectively change certain mind design attributes, temporarily producing a different mind.

**Size, Complexity and Properties of Minds**

Given that minds are countable they could be arranged in an ordered list, for example in order of numerical value of the representing string. This means that some mind will have the interesting property of being the smallest. If we accept that a Universal Turing Machine (UTM) is a type of mind, if we denote by ($m$, $n$) the class of UTMs with $m$ states and $n$ symbols, the following UTMs have been discovered: (9, 3), (4, 6), (5, 5), and (2, 18). The (4, 6)-UTM uses only 22 instructions, and no standard machine of lesser complexity has been found [8]. Alternatively, we may ask about the largest mind. Given that we have already shown that the set of minds is infinite, such an entity does not exist. However, if we take into account our embodiment requirement the largest mind may in fact correspond to the design at the physical limits of computation [9].

Another interesting property of the minds is that they all can be generated by a simple deterministic algorithm, a variant of Levin Search [10]: start with an integer (for example 42), check to see if the number encodes a mind, if not, we discard the number, otherwise we add it to the set of mind designs and proceed to examine the next integer. Every mind will eventually appear on our list of minds after a predetermined number of steps. However, checking to see if something is in fact a mind is not a trivial procedure. Rice's theorem [11] explicitly forbids determination of non-trivial properties of random programs. One way to overcome this limitation is to introduce an arbitrary time limit on the mind-or-not-mind determination function effectively avoiding the underlying halting problem.

Analyzing our mind-design generation algorithm we may raise the question of complexity measure for mind designs, not in terms of the abilities of the mind, but in terms of complexity of design representation. Our algorithm outputs minds in order of their increasing value, but this is not representative of the design complexity of the respective minds. Some minds may be represented by highly compressible numbers with a short representation such as $10^{13}$, while others may be comprised of 10,000 completely random digits, for example 7358348955651172160377535629140… [12]. We suggest that Kolmogorov Complexity (KC) [13] measure could be applied to strings representing mind designs. Consequently some minds will be rated as "elegant" – having a compressed representation much shorter than the original string while others will be "efficient" representing the most efficient representation of that particular mind. Interesting elegant minds might be easier to discover than efficient minds, but unfortunately KC is not generally computable.

In the context of complexity analysis of mind designs we can ask a few interesting philosophical questions. For example could two minds be added together [14], in other words, is it possible to combine two uploads or two artificially intelligent programs into a single, unified mind design? Could this process be reversed? Could a single mind be separated into multiple non-identical entities each in itself a mind? Additionally, could one mind design be changed into another via a gradual process without destroying it? For example could a computer virus (or even a real virus loaded with DNA of another person) be a sufficient cause to alter a mind into a predictable type of other mind? Could specific properties be introduced into a mind given this virus-based approach? For example could Friendliness [15] be added post factum to an existing mind design?

Each mind design corresponds to an integer and so is finite, but since the number of minds is infinite some have a much greater number of states compared to others. This property holds for all minds. Consequently, since a human mind has only a finite number of possible states, there are minds which can never be fully understood by a human mind as such mind designs have a much greater number of states, making their understanding impossible as can be demonstrated by the pigeonhole principle.

**Space of Mind Designs**

Overall the set of human minds (about 7 billion of them currently available and about 100 billion ever existed) is very homogeneous both in terms of hardware (embodiment in a human body) and software (brain design and knowledge). In fact the small differences between human minds are trivial in the context of the full infinite spectrum of possible mind designs. Human minds represent only a small constant size subset of the great mind landscape. Same could be said about the sets of other earthly minds such as dog minds, or bug minds or male minds or in general the set of all animal minds.

Given our algorithm for sequentially generating minds, one can see that a mind could never be completely destroyed, making minds theoretically immortal. A particular mind may not be embodied at a given time, but the idea of it is always present. In fact it was present even before the material universe came into existence. So, given sufficient computational resources any mind design could be regenerated, an idea commonly associated with the concept of reincarnation [16]. Also, the most powerful and most knowledgeable mind has always been associated with the idea of Deity or the Universal Mind.

Given our definition of mind we can classify minds with respect to their design, knowledgebase or embodiment. First, the designs could be classified with respect to their origins: copied from an existing mind like an upload, evolved via artificial or natural evolution or explicitly designed with a set of particular desirable properties. Another alternative is what is known as a Boltzmann Brain – a complete mind embedded in a system which arises due to statistically rare random fluctuations in the particles comprising the universe, but which is very likely due to vastness of cosmos [17].

Lastly a possibility remains that some minds are physically or informationally recursively nested within other minds. With respect to the physical nesting we can consider a type of mind suggested by Kelly [18] who talks about "a very slow invisible mind over large physical distances". It is possible that the physical universe as a whole or a significant part of it comprises such a mega-mind. That theory has been around for millennia and has recently received some indirect experimental support [19]. In that case all the other minds we can consider are nested within such larger mind. With respect to the informational nesting a powerful mind can generate a less powerful mind as an idea. This obviously would take some precise thinking but should be possible for a sufficiently powerful artificially intelligent mind. Some scenarios describing informationally nested minds are analyzed by Yampolskiy in his work on artificial intelligence confinement problem [7]. Bostrom, using statistical reasoning, suggests that all observed minds, and the whole universe, are nested within a mind of a very powerful computer [20]. Similarly Lanza, using a completely different and somewhat controversial approach (biocentrism), argues that the universe is created by biological minds [21]. It remains to be seen if given a particular mind its origins can be deduced from some detailed analysis of the minds design or actions.

While minds designed by human engineers comprise only a tiny region in the map of mind designs it is probably the best explored part of the map. Numerous surveys of artificial minds, created by AI researchers in the last 50 years, have been produced [22-26]. Such surveys typically attempt to analyze state-of-the-art in artificial cognitive systems and provide some internal classification of dozens of the reviewed systems with regards to their components and overall design. The main subcategories into which artificial minds designed by human engineers can be placed include brain (at the neuron level) emulators [24], biologically inspired cognitive architectures [25], physical symbol systems, emergent systems, dynamical and enactive systems [26]. Rehashing information about specific architectures presented in such surveys is beyond the scope of this paper, but one can notice incredible richness and diversity of designs even in that tiny area of the overall map we are trying to envision. For readers particularly interested in overview of superintelligent minds, animal minds and possible minds in addition to surveys mentioned above a recent paper "Artificial General Intelligence and the Human Mental Model" by Yampolskiy and Fox is highly recommended [27].

For each mind subtype there are numerous architectures, which to a certain degree depend on the computational resources available via a particular embodiment. For example, theoretically a mind working with infinite computational resources could trivially brute-force any problem, always arriving at the optimal solution, regardless of its size. In practice, limitations of the physical world place constraints on available computational resources regardless of the embodiment type, making brute-force approach a non-feasible solution for most real world problems [9]. Minds working with limited computational resources have to rely on heuristic simplifications to arrive at "good enough" solutions [28-31].

Another subset of architectures consists of self-improving minds. Such minds are capable of examining their own design and finding improvements in their embodiment, algorithms or knowledgebases which will allow the mind to more efficiently perform desired operations [32]. It is very likely that possible improvements would form a Bell curve with many initial opportunities for optimization towards higher efficiency and fewer such options remaining after every generation. Depending on the definitions used, one can argue that a recursively self-improving mind actually changes itself into a different mind, rather than remaining itself, which is particularly obvious after a sequence of such improvements. Taken to extreme this idea implies that a simple act of learning new information transforms you into a different mind raising millennia old questions about the nature of personal identity.

With respect to their knowledgebases minds could be separated into those without an initial knowledgebase, and which are expected to acquire their knowledge from the environment, minds which are given a large set of universal knowledge from the inception and those minds which are given specialized knowledge only in one or more domains. Whether the knowledge is stored in an efficient manner, compressed, classified or censored is dependent on the architecture and is a potential subject of improvement by self-modifying minds.

One can also classify minds in terms of their abilities or intelligence. Of course the problem of measuring intelligence is that no universal tests exist. Measures such as IQ tests and performance on specific tasks are not universally accepted and are always highly biased against non-human intelligences. Recently some work has been done on streamlining intelligence measurements across different types of machine intelligence [2, 33] and other "types" of intelligence [34], but the applicability of the results is still being debated. In general, the notion of intelligence only makes sense in the context of problems to which said intelligence can be applied. In fact this is exactly how IQ tests work, by presenting the subject with a number of problems and seeing how many the subject is able to solve in a given amount of time (computational resource). A subfield of computer science known as computational complexity theory is devoted to studying and classifying different problems with respect to their difficulty and with respect to computational resources necessary to solve them. For every class of problems complexity theory defines a class of machines capable of solving such problems. We can apply similar ideas to classifying minds, for example all minds capable of efficiently [12] solving problems in the class P or a more difficult class of NP-complete problems [35]. Similarly we can talk about minds with general intelligence belonging to the class of AI-Complete [36-38] minds, such as humans.

We can also look at the goals of different minds. It is possible to create a system which has no terminal goals and so such a mind is not very motivated to accomplish things. Many minds are designed or trained for obtaining a particular high level goal or a set of goals. We can envision a mind which has randomly changing goal or a set of goals, as well as a mind which has many goals of different priority. Steve Omohundro used micro-economic theory to speculate about the driving forces in the behavior of superintelligent machines. He argues that intelligent machines will want to self-improve, be rational, preserve their utility functions, prevent counterfeit utility [39], acquire resources and use them efficiently, and protect themselves. He believes that machines' actions will be governed by rational economic behavior [40, 41]. Mark Waser suggested an additional "drive" to be included in the list of behaviors predicted to be exhibited by the machines [42]. Namely, he suggests that evolved desires for cooperation and being social are part of human ethics and are a great way of accomplishing goals, an idea also analyzed by Joshua Fox and Carl Shulman, but with contrary conclusions [43]. While it is commonly assumed that minds with high intelligence will converge on a common goal, Nick Bostrom via his orthogonality thesis has argued that a system can have any combination of intelligence and goals [44].

Regardless of design, embodiment or any other properties, all minds can be classified with respect to two fundamental but scientifically poorly defined properties – free will and consciousness. Both descriptors suffer from an ongoing debate regarding their actual existence or explanatory usefulness. This is primarily a result of impossibility to design a definitive test to measure or even detect said properties, despite numerous attempts [45-47] or to show that theories associated with them are somehow falsifiable. Intuitively we can speculate that consciousness, and maybe free will, are not binary properties but rather continuous and emergent abilities commensurate with a degree of general intelligence possessed by the system or some other property we shall term "mindness". Free will can be said to correlate with a degree to which behavior of the system can't be predicted [48]. This is particularly important in the design of artificially intelligent systems for which inability to predict their future behavior is a highly undesirable property from the safety point of view [49, 50]. Consciousness on the other hand seems to have no important impact on the behavior of the system as can be seen from some thought experiments supposing existence of "consciousless" intelligent agents [51]. This may change if we are successful in designing a test, perhaps based on observer impact on quantum systems [52], to detect and measure consciousness.

In order to be social, two minds need to be able to communicate which might be difficult if the two minds don't share a common communication protocol, common culture or even common environment. In other words, if they have no common grounding they don't understand each other. We can say that two minds understand each other if given the same set of inputs they produce similar outputs. For example, in sequence prediction tasks [53] two minds have an understanding if their predictions are the same regarding the future numbers of the sequence based on the same observed subsequence. We can say that a mind can understand another mind's function if it can predict the other's output with high accuracy. Interestingly, a perfect ability by two minds to predict each other would imply that they are identical and that they have no free will as defined above.

**A Survey of Taxonomies**

Yudkowsky describes the map of mind design space as follows: "In one corner, a tiny little circle contains all humans; within a larger tiny circle containing all biological life; and all the rest of the huge map is the space of minds-in-general. The entire map floats in a still vaster space, the space of optimization processes. Natural selection creates complex functional machinery without mindfulness; evolution lies inside the space of optimization processes but outside the circle of minds" [54]. Figure 1 illustrates one possible mapping inspired by this description.

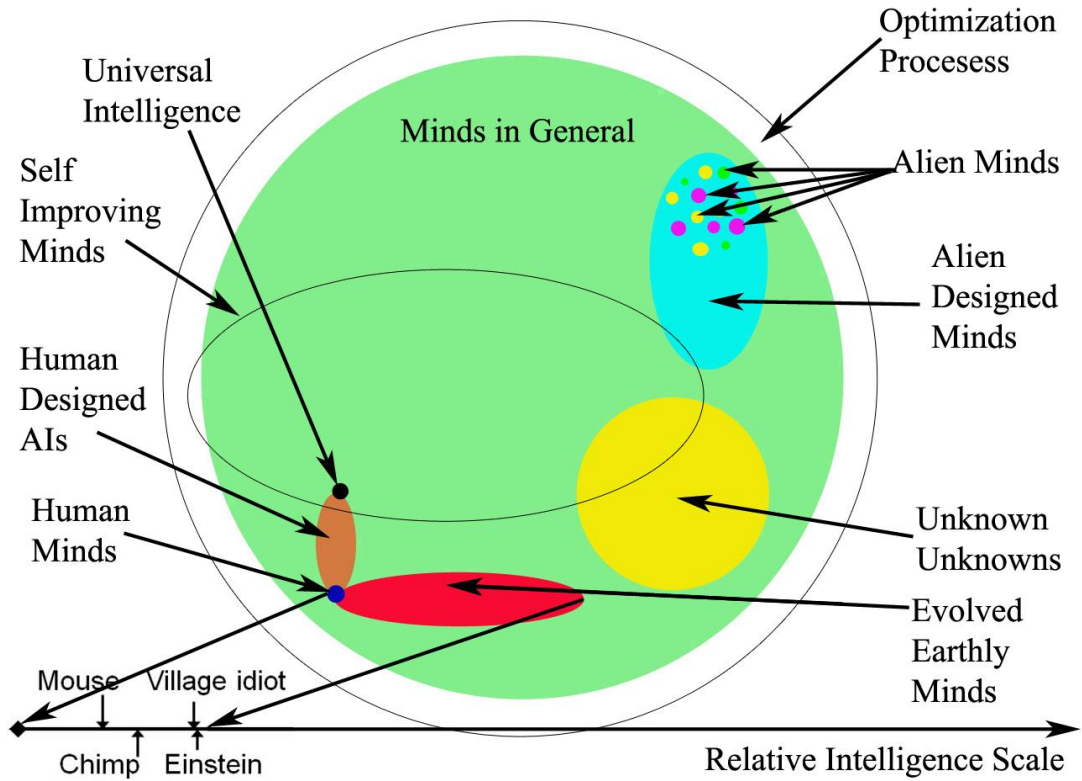

Figure 1: The universe of possible minds [54, 55].

Similarly, Ivan Havel writes "… all conceivable cases of intelligence (of people, machines, whatever) are represented by points in a certain abstract multi-dimensional "super space" that I will call the intelligence space (shortly IS). Imagine that a specific coordinate axis in IS is assigned to any conceivable particular ability, whether human, machine, shared, or unknown (all axes having one common origin). If the ability is measurable the assigned axis is endowed with a corresponding scale. Hypothetically, we can also assign scalar axes to abilities, for which only relations like "weaker-stronger", "better-worse", "less-more" etc. are meaningful; finally, abilities that may be only present or absent may be assigned with "axes" of two (logical) values (yes-no). Let us assume that all coordinate axes are oriented in such a way that greater distance from the common origin always corresponds to larger extent, higher grade, or at least to the presence of the corresponding ability. The idea is that for each individual intelligence (i.e. the intelligence of a particular person, machine, network, etc.), as well as for each generic intelligence (of some group) there exists just one representing point in IS, whose coordinates determine the extent of involvement of particular abilities [56]." If the universe (or multiverse) is infinite, as our current physics theories indicate, then all possible minds in all possible states are instantiated somewhere [4].

Ben Goertzel proposes the following classification of Kinds of Minds, mostly centered around the concept of embodiment [57]:

- **Singly Embodied** – control a single physical or simulated system.
- **Multiply Embodied** - control a number of disconnected physical or simulated systems.
- **Flexibly Embodied** – control a changing number of physical or simulated systems.
- **Non-Embodied** – resides in a physical substrate but doesn't utilize the body in a traditional way.
- **Body-Centered** – consists of patterns emergent between physical system and the environment.
- **Mindplex** – a set of collaborating units each of which is itself a mind [58].
- **Quantum** – an embodiment based on properties of quantum physics.
- **Classical** - an embodiment based on properties of classical physics.

J. Storrs Hall in his "Kinds of Minds" suggests that different stages a developing AI may belong to can be classified relative to its humanlike abilities. His classification encompasses:

- **Hypohuman** - infrahuman, less-than-human capacity.
- **Diahuman** - human-level capacities in some areas, but still not a general intelligence.
- **Parahuman** - similar but not identical to humans, as for example, augmented humans.
- **Allohuman** - as capable as humans, but in different areas.
- **Epihuman** - slightly beyond the human level.
- **Hyperhuman** - much more powerful than human, superintelligent [27, 59].

Patrick Roberts in his book *Mind Making* presents his ideas for a "Taxonomy of Minds", we will leave it to the reader to judge usefulness of his classification [60]:

- **Choose Means** - Does it have redundant means to the same ends? How well does it move between them?

- **Mutate** - Can a mind naturally gain and lose new ideas in its lifetime?
- **Doubt** - Is it eventually free to lose some or all beliefs? Or is it wired to obey the implications of every sensation?
- **Sense Itself** - Does a mind have the senses to see the physical conditions of that mind?
- **Preserve Itself** - Does a mind also have the means to preserve or reproduce itself?
- **Sense Minds** - Does a mind understand mind, at least of lower classes, and how well does it apply that to itself, to others?
- **Sense Kin** - Can it recognize the redundant minds, or at least the bodies of minds, that it was designed to cooperate with?
- **Learn** - Does the mind's behavior change from experience? Does it learn associations?
- **Feel** - We imagine that an equally intelligent machine would lack our conscious experience.
- **Communicate** - Can it share beliefs with other minds?

Kevin Kelly has also proposed a "Taxonomy of Minds" which in his implementation is really just a list of different minds, some of which have not showed up in other taxonomies [18]:

- "Super fast human mind.
- Mind with operational access to its source code.
- Any mind capable of general intelligence and self-awareness.
- General intelligence without self-awareness.
- Self-awareness without general intelligence.
- Super logic machine without emotion.
- Mind capable of imagining greater mind.
- Mind capable of creating greater mind. (M2)
- Self-aware mind incapable of creating a greater mind.
- Mind capable of creating greater mind which creates greater mind. etc. (M3, and Mn)
- Mind requiring protector while it develops.
- Very slow "invisible" mind over large physical distance.
- Mind capable of cloning itself and remaining in unity with clones.
- Mind capable of immortality.
- Rapid dynamic mind able to change its mind-space-type sectors (think different)
- Global mind -- large supercritical mind of subcritical brains.
- Hive mind -- large super critical mind made of smaller minds each of which is supercritical.
- Low count hive mind with few critical minds making it up.
- Borg -- supercritical mind of smaller minds supercritical but not self-aware
- Nano mind -- smallest (size and energy profile) possible super critical mind.
- Storebit -- Mind based primarily on vast storage and memory.
- Anticipators -- Minds specializing in scenario and prediction making.
- Guardian angels -- Minds trained and dedicated to enhancing your mind, useless to anyone else.
- Mind with communication access to all known "facts." (F1)
- Mind which retains all known "facts," never erasing. (F2)
- Symbiont, half machine half animal mind.

- Cyborg, half human half machine mind.
- Q-mind, using quantum computing
- Vast mind employing faster-than-light communications"

Elsewhere Kelly provides a lot of relevant analysis of landscape of minds writing about Inevitable Minds [61], The Landscape of Possible Intelligences [62], What comes After Minds? [63], and the Evolutionary Mind of God [64].

Aaron Sloman in "The Structure of the Space of Possible Minds", using his virtual machine model, proposes a division of the space of possible minds with respect to the following properties [1]:

- Quantitative VS Structural
- Continuous VS Discrete
- Complexity of stored instructions
- Serial VS Parallel
- Distributed VS Fundamentally Parallel
- Connected to External Environment VS Not Connected
- Moving VS Stationary
- Capable of modeling others VS Not capable
- Capable of logical inference VS Not Capable
- Fixed VS Re-programmable
- Goal consistency VS Goal Selection
- Meta-Motives VS Motives
- Able to delay goals VS Immediate goal following
- Statics Plan VS Dynamic Plan
- Self-aware VS Not Self-Aware

**Mind Cloning and Equivalence Testing Across Substrates**

The possibility of uploads rests on the ideas of computationalism [65] specifically, substrate independence and equivalence meaning that the same mind can be instantiated in different substrates and move freely between them. If your mind is cloned and if a copy is instantiated in a different substrate from the original one (or on the same substrate), how can it be verified that the copy is indeed an identical mind? At least immediately after cloning and before it learns any new information. For that purpose I propose a variant of a Turing Test, which also relies on interactive text-only communication to ascertain quality of the copied mind. The text-only interface is important not to prejudice the examiner against any unusual substrates on which the copied mind might be running. The test proceeds by having the examiner (original mind) ask questions of the copy (cloned mind), questions which supposedly only the original mind would know answers to (testing should be done in a way which preserves privacy). Good questions would relate to personal preferences, secrets (passwords, etc.) as well as recent dreams. Such test could also indirectly test for consciousness via similarity of subjective qualia. Only a perfect copy should be able to answers all such questions in the same way as the original mind. Another variant of the same test may have a 3rd party test the original and cloned mind by seeing if they always provide the same answer to any question. One needs to be careful in such questioning not to give undue weight to questions related to the minds substrate as that may lead to different answers. For example, asking a human if he is hungry may produce an answer different from the one which would be given by a non-biological robot.

**Conclusions**

Science periodically experiences a discovery of a whole new area of investigation. For example, observations made by Galileo Galilei lead to the birth of observational astronomy [66], aka study of our universe; Watson and Crick's discovery of the structure of DNA lead to the birth of the field of genetics [67], which studies the universe of blueprints for organisms; Stephen Wolfram's work with cellular automata has resulted in "a new kind of science" [68] which investigates the universe of computational processes. I believe that we are about to discover yet another universe – the universe of minds.

As our understanding of human brain improves, thanks to numerous projects aimed at simulating or reverse engineering a human brain, we will no doubt realize that human intelligence is just a single point in the vast universe of potential intelligent agents comprising a new area of study. The new field, which I would like to term *intellectology*, will study and classify design space of intelligent agents, work on establishing limits to intelligence (minimum sufficient for general intelligence and maximum subject to physical limits), contribute to consistent measurement of intelligence across intelligent agents, look at recursive self-improving systems, design new intelligences (making AI a sub-field of intellectology) and evaluate capacity for understanding higher level intelligences by lower level ones. At the more theoretical level the field will look at the distribution of minds on the number line and probabilistic distribution of minds in the mind design space as well as attractors in the mind design space. It will consider how evolution, drives and design choices impact density of minds in the space of possibilities. It will investigate intelligence as an additional computational resource along time and memory. The field will not be subject to the current limitations brought on by the human centric view of intelligence and will open our understanding to seeing intelligence as a fundamental resource like space or time. Finally, I believe intellectology will highlight inhumanity of most possible minds and the dangers associated with such minds being placed in charge of humanity.

References

1. Sloman, A., *The Structure and Space of Possible Minds*. The Mind and the Machine: philosophical aspects of Artificial Intelligence1984: Ellis Horwood LTD.
2. Legg, S. and M. Hutter, *Universal Intelligence: A Definition of Machine Intelligence.* Minds and Machines, December 2007. **17(4)**: p. 391-444.
3. Hanson, R., *If Uploads Come First.* Extropy, 1994. **6(2)**.
4. Bostrom, N., *Quantity of experience: brain-duplication and degrees of consciousness.* Minds and Machines, 2006. **16(2)**: p. 185-200.
5. Yampolskiy, R. and M. Gavrilova, *Artimetrics: Biometrics for Artificial Entities.* IEEE Robotics and Automation Magazine (RAM), 2012. **19**(4): p. 48-58.

6. Yampolskiy, R.V., B. Klare, and A.K. Jain. *Face recognition in the virtual world: Recognizing Avatar faces*. in *Machine Learning and Applications (ICMLA), 2012 11th International Conference on*. 2012. IEEE.
7. Yampolskiy, R.V., *Leakproofing Singularity - Artificial Intelligence Confinement Problem.* Journal of Consciousness Studies (JCS), 2012. **19(1-2)**: p. 194–214.
8. Wikipedia, *Universal Turing Machine*, Retrieved April 14, 2011: Available at: http://en.wikipedia.org/wiki/Universal_Turing_machine.
9. Lloyd, S., *Ultimate Physical Limits to Computation.* Nature, 2000. **406**: p. 1047-1054.
10. Levin, L., *Universal Search Problems.* Problems of Information Transmission, 1973. **9(3)**: p. 265--266.
11. Rice, H.G., *Classes of recursively enumerable sets and their decision problems.* Transactions of the American Mathematical Society, 1953. **74**(2): p. 358-366.
12. Yampolskiy, R.V., *Efficiency Theory: a Unifying Theory for Information, Computation and Intelligence.* Journal of Discrete Mathematical Sciences & Cryptography, 2013. **16(4-5)**: p. 259-277.
13. Kolmogorov, A.N., *Three Approaches to the Quantitative Definition of Information.* Problems Inform. Transmission, 1965. **1(1)**: p. 1-7.
14. Sotala, K. and H. Valpola, *Coalescing Minds: Brain Uploading-Related Group Mind Scenarios.* International Journal of Machine Consciousness, 2012. **4(1)**: p. 293-312.
15. Yudkowsky, E.S., *Creating Friendly AI - The Analysis and Design of Benevolent Goal Architectures*, 2001: Available at: http://singinst.org/upload/CFAI.html.
16. Fredkin, E., *On the soul*, 1982, Draft.
17. De Simone, A., et al., *Boltzmann brains and the scale-factor cutoff measure of the multiverse.* Physical Review D, 2010. **82**(6): p. 063520.
18. Kelly, K., *A Taxonomy of Minds*, 2007: Available at: http://kk.org/thetechnium/archives/2007/02/a_taxonomy_of_m.php.
19. Krioukov, D., et al., *Network Cosmology.* Sci. Rep., 2012. **2**.
20. Bostrom, N., *Are You Living In a Computer Simulation?* Philosophical Quarterly, 2003. **53(211)**: p. 243-255.
21. Lanza, R., *A new theory of the universe.* American Scholar, 2007. **76**(2): p. 18.
22. Miller, M.S.P. *Patterns for Cognitive Systems*. in *Complex, Intelligent and Software Intensive Systems (CISIS), 2012 Sixth International Conference on*. 2012.
23. Cattell, R. and A. Parker, *Challenges for Brain Emulation: Why is it so Difficult?* Natural Intelligence, 2012. **1(3)**: p. 17-31.
24. de Garis, H., et al., *A world survey of artificial brain projects, Part I: Large-scale brain simulations.* Neurocomputing, 2010. **74**(1–3): p. 3-29.
25. Goertzel, B., et al., *A world survey of artificial brain projects, Part II: Biologically inspired cognitive architectures.* Neurocomput., 2010. **74**(1-3): p. 30-49.
26. Vernon, D., G. Metta, and G. Sandini, *A Survey of Artificial Cognitive Systems: Implications for the Autonomous Development of Mental Capabilities in Computational Agents.* IEEE Transactions on Evolutionary Computation, 2007. **11**(2): p. 151-180.
27. Yampolskiy, R.V. and J. Fox, *Artificial General Intelligence and the Human Mental Model*, in *Singularity Hypotheses*2012, Springer Berlin Heidelberg. p. 129-145.
28. Yampolskiy, R.V., L. Ashby, and L. Hassan, *Wisdom of Artificial Crowds—A Metaheuristic Algorithm for Optimization.* Journal of Intelligent Learning Systems and Applications, 2012. **4**(2): p. 98-107.

29. Ashby, L.H. and R.V. Yampolskiy. *Genetic algorithm and Wisdom of Artificial Crowds algorithm applied to Light up*. in *Computer Games (CGAMES), 2011 16th International Conference on*. 2011. IEEE.
30. Hughes, R. and R.V. Yampolskiy, *Solving Sudoku Puzzles with Wisdom of Artificial Crowds.* International Journal of Intelligent Games & Simulation, 2013. **7**(1): p. 6.
31. Port, A.C. and R.V. Yampolskiy. *Using a GA and Wisdom of Artificial Crowds to solve solitaire battleship puzzles*. in *Computer Games (CGAMES), 2012 17th International Conference on*. 2012. IEEE.
32. Hall, J.S., *Self-Improving AI: An Analysis.* Minds and Machines, October 2007. **17(3)**: p. 249 - 259.
33. Yonck, R., *Toward a Standard Metric of Machine Intelligence.* World Future Review, 2012. **4**(2): p. 61-70.
34. Herzing, D.L., *Profiling nonhuman intelligence: An exercise in developing unbiased tools for describing other "types" of intelligence on earth.* Acta Astronautica, 2014. **94**(2): p. 676-680.
35. Yampolskiy, R.V., *Construction of an NP Problem with an Exponential Lower Bound.* Arxiv preprint arXiv:1111.0305, 2011.
36. Yampolskiy, R.V., *Turing Test as a Defining Feature of AI-Completeness*, in *Artificial Intelligence, Evolutionary Computation and Metaheuristics - In the footsteps of Alan Turing. Xin-She Yang (Ed.)*2013, Springer. p. 3-17.
37. Yampolskiy, R.V., *AI-Complete, AI-Hard, or AI-Easy–Classification of Problems in AI.* The 23rd Midwest Artificial Intelligence and Cognitive Science Conference, Cincinnati, OH, USA, 2012.
38. Yampolskiy, R.V., *AI-Complete CAPTCHAs as Zero Knowledge Proofs of Access to an Artificially Intelligent System.* ISRN Artificial Intelligence, 2011. **271878**.
39. Yampolskiy, R.V., *Utility Function Security in Artificially Intelligent Agents.* Journal of Experimental and Theoretical Artificial Intelligence (JETAI), 2014: p. 1-17.
40. Omohundro, S.M., *The Nature of Self-Improving Artificial Intelligence*, in *Singularity Summit*2007: San Francisco, CA.
41. Omohundro, S.M., *The Basic AI Drives*, in *Proceedings of the First AGI Conference, Volume 171, Frontiers in Artificial Intelligence and Applications, P. Wang, B. Goertzel, and S. Franklin (eds.)*February 2008, IOS Press.
42. Waser, M.R., *Designing a Safe Motivational System for Intelligent Machines*, in *The Third Conference on Artificial General Intelligence*March 5-8, 2010: Lugano, Switzerland.
43. Fox, J. and C. Shulman, *Superintelligence Does Not Imply Benevolence*, in *8th European Conference on Computing and Philosophy*October 4-6, 2010 Munich, Germany.
44. Bostrom, N., *The superintelligent will: Motivation and instrumental rationality in advanced artificial agents.* Minds and Machines, 2012. **22**(2): p. 71-85.
45. Hales, C., *An empirical framework for objective testing for P-consciousness in an artificial agent.* Open Artificial Intelligence Journal, 2009. **3**: p. 1-15.
46. Aleksander, I. and B. Dunmall, *Axioms and Tests for the Presence of Minimal Consciousness in Agents I: Preamble.* Journal of Consciousness Studies, 2003. **10**(4-5): p. 4-5.
47. Arrabales, R., A. Ledezma, and A. Sanchis, *ConsScale: a plausible test for machine consciousness?* 2008.

48. Aaronson, S., *The Ghost in the Quantum Turing Machine.* arXiv preprint arXiv:1306.0159, 2013.
49. Yampolskiy, R.V., *Artificial intelligence safety engineering: Why machine ethics is a wrong approach*, in *Philosophy and Theory of Artificial Intelligence*2013, Springer Berlin Heidelberg. p. 389-396.
50. Yampolskiy, R.V., *What to Do with the Singularity Paradox?*, in *Philosophy and Theory of Artificial Intelligence*2013, Springer Berlin Heidelberg. p. 397-413.
51. Chalmers, D.J., *The conscious mind: In search of a fundamental theory*1996: Oxford University Press.
52. Gao, S., *A quantum method to test the existence of consciousness.* The Noetic Journal, 2002. **3**(3): p. 27-31.
53. Legg, S. *Is there an elegant universal theory of prediction?* in *Algorithmic Learning Theory*. 2006. Springer.
54. Yudkowsky, E., *Artificial Intelligence as a Positive and Negative Factor in Global Risk*, in *Global Catastrophic Risks*, N. Bostrom and M.M. Cirkovic, Editors. 2008, Oxford University Press: Oxford, UK. p. 308-345.
55. Yudkowsky, E., *The Human Importance of the Intelligence Explosion*, in *Singularity Summit at Stanford*2006.
56. Havel, I.M., *On the Way to Intelligence Singularity*, in *Beyond Artificial Intelligence*, J. Kelemen, J. Romportl, and E. Zackova, Editors. 2013, Springer Berlin Heidelberg. p. 3-26.
57. Geortzel, B., *The Hidden Pattern: A Patternist Philosophy of Mind. Chapter 2 - Kinds of Minds* 2006: Brown Walker Press.
58. Goertzel, B., *Mindplexes: The Potential Emergence of Multiple Levels of Focused Consciousness in Communities of AI's and Humans* Dynamical Psychology, 2003. **http://www.goertzel.org/dynapsyc/2003/mindplex.htm**.
59. Hall, J.S., *Chapter 15: Kinds of Minds*, in *Beyond AI: Creating the Conscience of the Machine*2007, Prometheus Books: Amherst, NY.
60. Roberts, P., *Mind Making: The Shared Laws of Natural and Artificial*2009: CreateSpace.
61. Kelly, K., *Inevitable Minds*, 2009: Available at: http://kk.org/thetechnium/archives/2009/04/inevitable_mind.php.
62. Kelly, K., *The Landscape of Possible Intelligences*, 2008: Available at: http://kk.org/thetechnium/archives/2008/09/the_landscape_o.php.
63. Kelly, K., *What Comes After Minds?*, 2008: Available at: http://kk.org/thetechnium/archives/2008/12/what_comes_afte.php.
64. Kelly, K., *The Evolutionary Mind of God* 2007: Available at: http://kk.org/thetechnium/archives/2007/02/the_evolutionar.php.
65. Putnam, H., *Brains and behavior.* Readings in philosophy of psychology, 1980. **1**: p. 24-36.
66. Galilei, G., *Dialogue concerning the two chief world systems: Ptolemaic and Copernican*1953: University of California Pr.
67. Watson, J.D. and F.H. Crick, *Molecular structure of nucleic acids.* Nature, 1953. **171**(4356): p. 737-738.
68. Wolfram, S., *A New Kind of Science*May 14, 2002: Wolfram Media, Inc.